%% file: main.tex
\documentclass[conference]{IEEEtran}
\usepackage{times}
\usepackage{multirow}
\usepackage[numbers]{natbib}
\usepackage{multicol}
\usepackage[bookmarks=true]{hyperref}
\usepackage{tabularx}
\usepackage{booktabs}
\input{math_commands.tex}

\usepackage{graphicx}
\usepackage{algorithm}
\usepackage{algorithmic}

\usepackage{hyperref}
\usepackage{url}
\usepackage{tcolorbox}
\usepackage{graphicx}
\usepackage{caption}
\usepackage{lipsum}
\usepackage{stfloats}
\usepackage{subfigure}
\hypersetup{
    pdfborder={0 0 0},   % No border
    colorlinks=true,      % Colored links (you can also use `false` if you don't want any color)
   linkcolor=blue,  % Custom dark blue for internal links
    urlcolor=blue    % Custom dark blue for external links
}

\newcommand{\mnamefull}{DexterityGen}
\newcommand{\mname}{DexGen}
\title{
\textsc{\mnamefull}:\\Foundation Controller for Unprecedented Dexterity
}

\author{Zhao-Heng Yin$^{1,2}$, Changhao Wang$^{2}$, Luis Pineda$^{2}$, Francois Hogan$^{2}$, Krishna Bodduluri$^{2}$, Akash Sharma$^{2}$, \\ 
Patrick Lancaster$^{2}$, Ishita Prasad$^{2}$, Mrinal Kalakrishnan$^{2}$, Jitendra Malik$^{2}$, Mike Lambeta$^{2}$, \\ Tingfan Wu$^{2}$, Pieter Abbeel$^{1}$, Mustafa Mukadam$^{2}$\\ \\ 
$^1$BAIR, UC Berkeley\quad $^2$FAIR at Meta\\
{\href{https://zhaohengyin.github.io/dexteritygen}{\texttt{zhaohengyin.github.io/dexteritygen}}}
\vspace{-0.3cm}
}

\begin{document}

\twocolumn[{%
\renewcommand\twocolumn[1][]{#1}%

\maketitle
\begin{center}
    \includegraphics[width=0.995\linewidth]{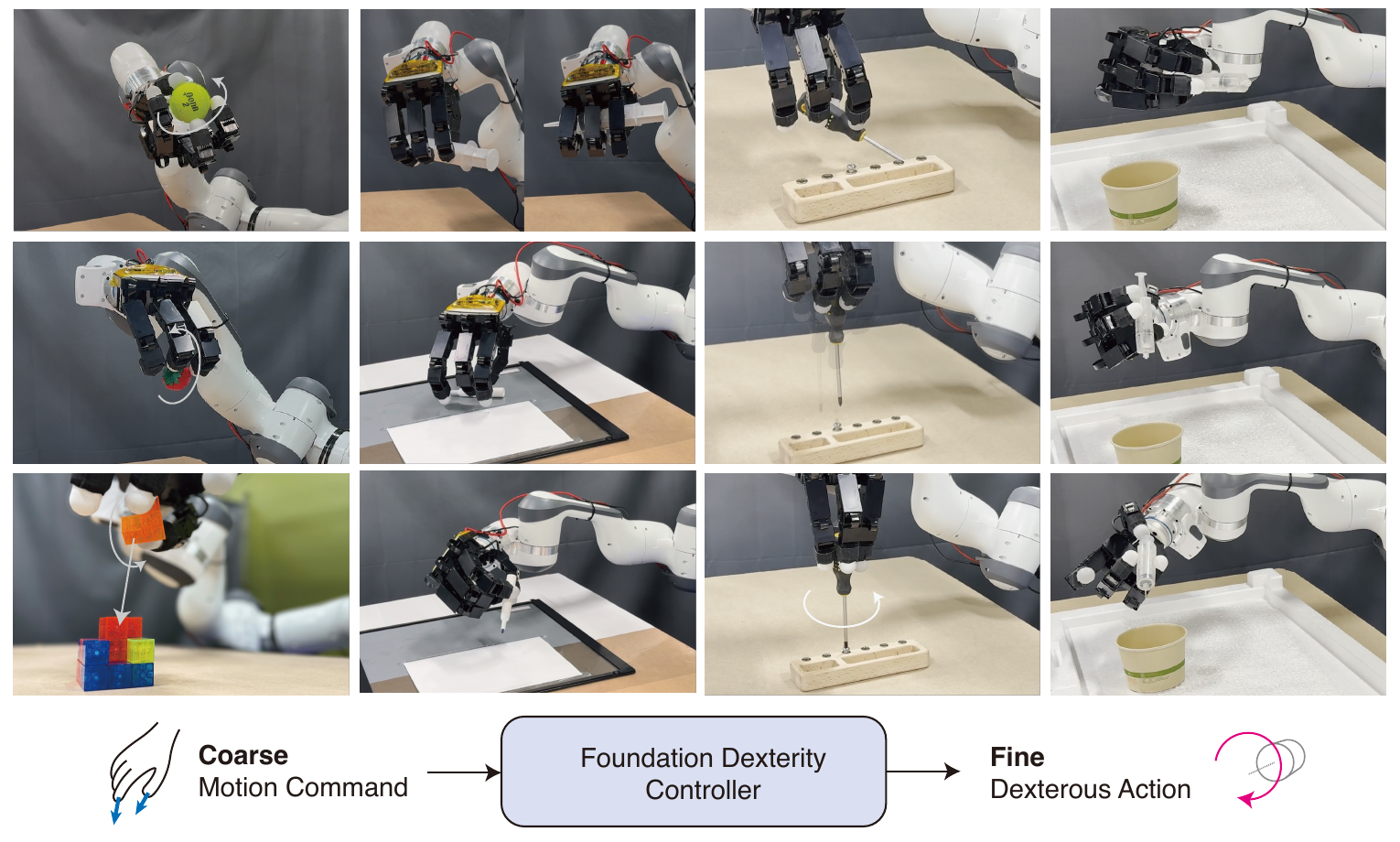}

    \captionof{figure}{We introduce \mnamefull{}~(\mname{}) as a foundation controller that achieves unprecedented dexterous manipulation behavior with teleoperation. \mname{} is a generative model that can translate an unsafe, coarse motion command produced by external policy to safe and fine actions. With human teleoperation as a high-level policy, \mname{} exhibits unprecedented dexterity from diverse object rotation and regrasping to using pen, syringe, and screwdriver.}
    \label{fig:teaser}
   \vspace{-0.1cm}
  
\end{center}
}]

\begin{abstract}
Teaching robots dexterous manipulation skills, such as tool use, presents a significant challenge.  Current approaches can be broadly categorized into two strategies: human teleoperation (for imitation learning) and sim-to-real reinforcement learning. The first approach is difficult as it is hard for humans to produce safe and dexterous motions on a different embodiment without touch feedback. The second RL-based approach struggles with the domain gap and involves highly task-specific reward engineering on complex tasks. Our key insight is that RL is effective at learning low-level motion primitives, while humans excel at providing coarse motion commands for complex, long-horizon tasks. Therefore, the optimal solution might be a combination of both approaches. In this paper, we introduce \mnamefull{}~(\mname{}), which uses RL to pretrain large-scale dexterous motion primitives, such as in-hand rotation or translation. We then leverage this learned dataset to train a dexterous foundational controller. In the real world, we use human teleoperation as a prompt to the controller to produce highly dexterous behavior. We evaluate the effectiveness of \mname{} in both simulation and real world, demonstrating that it is a general-purpose controller that can realize input dexterous manipulation commands and significantly improves stability by 10-100x measured as duration of holding objects across diverse tasks. Notably, with \mname{} we demonstrate unprecedented dexterous skills including diverse object reorientation and dexterous tool use such as pen, syringe, and screwdriver for the first time.

\end{abstract}

\input{src/introduction}

\input{src/background}
\input{src/method}
\input{src/experiment}

\input{src/related_work}
\input{src/conclusion}

\section*{Acknowledgments}
This work was partially carried out during Zhao-Heng Yin's intern at the Meta FAIR Labs. This work is supported by the Meta FAIR Labs and ONR MURI N00014-22-1-2773. Pieter Abbeel holds concurrent appointments as a Professor at UC Berkeley and as an Amazon Scholar. This paper describes work performed at UC Berkeley and is not associated with Amazon. 

\bibliographystyle{plainnat}
\bibliography{references}
% \ \\ \\
% \newpage
\appendix

\input{src/appendix}

\end{document}

%% file: math_commands.tex
\usepackage{amsmath,amsfonts,bm}

\def\eqref#1{equation~\ref{#1}}
\def\1{\bm{1}}

\DeclareMathAlphabet{\mathsfit}{\encodingdefault}{\sfdefault}{m}{sl}
\SetMathAlphabet{\mathsfit}{bold}{\encodingdefault}{\sfdefault}{bx}{n}

%% file: src/introduction.tex
\begin{figure*}
    \centering
    \includegraphics[width=1.0\linewidth]{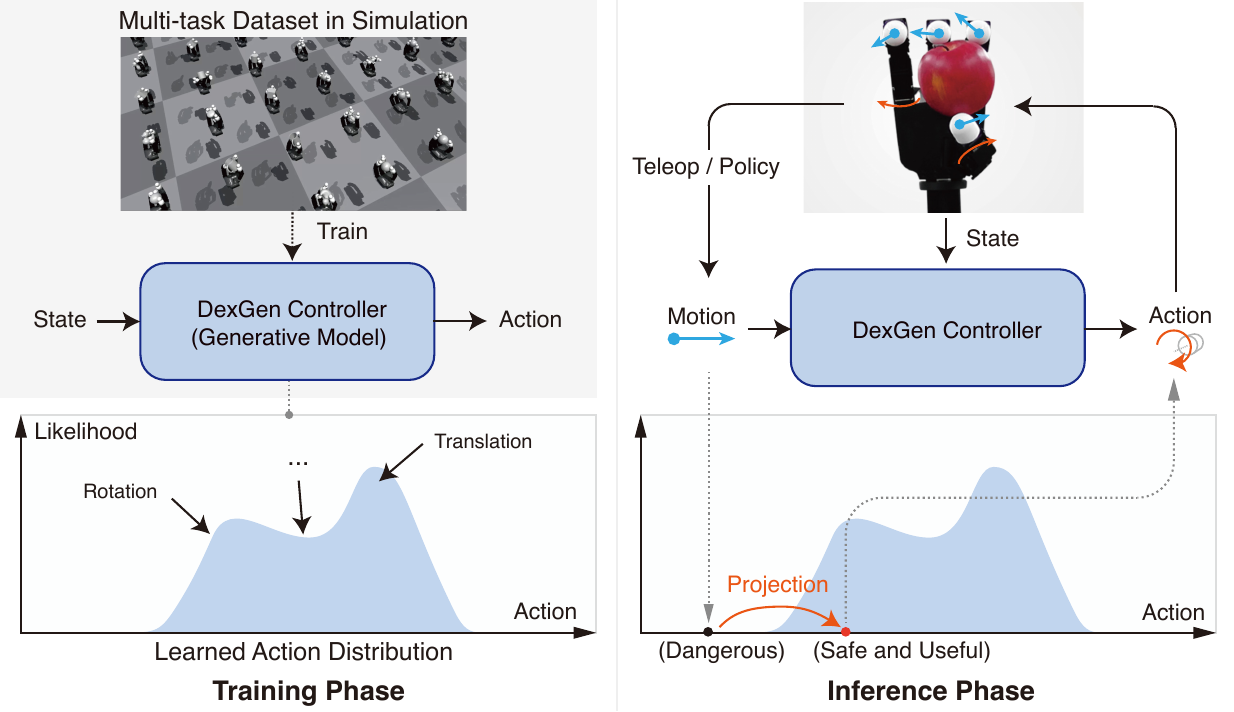}
    \caption{Overview of proposed framework. \textbf{Left (Training):} We collect a large multi-task dexterous in-hand manipulation dataset in simulation to pretrain a generative model that can generate diverse actions conditioned on the current state. The pretrained generative model can produce useful actions including rotation, translation, and more intricated behaviors.  \textbf{Right (Inference):} During inference, we can project dangerous motion produced by teleoperation or policy back to a high-likelihood action with guided sampling. This makes \mname{} capable of assisting a coarse high-level policy to perform complex object manipulations.}
    \label{fig:enter-label}
\end{figure*}

\section{Introduction}
Dexterous robotic hands are increasingly capturing attention due to their potential across various fields, including manufacturing, household tasks, and healthcare~\cite{okamura2000overview}. These robotic systems can replicate the fine motor skills of the human hand, enabling complex object manipulation~\cite{shaw2023leap, andrychowicz2020learning}. Their ability to perform tasks requiring human-like dexterity makes them valuable in areas where traditional automation falls short. However, effectively teaching dexterous in-hand manipulation skills to robotic hands remains a key challenge in robotics.

Recent data-driven approaches to teach robots dexterous manipulation skills can be boardly categorized into two categories: human teleoperation (for imitation learning)~\cite{hussein2017imitation, handa2020dexpilot,sivakumar2022robotic,qin2023anyteleop,ding2024bunny,lin2024learning,cheng2024open,wang2024dexcap,shaw2024bimanual} and sim-to-real Reinforcement Learning~(RL)~\cite{andrychowicz2020learning,handa2023dextreme,qi2023hand,yin2023rotating,khandate2023sampling,huang2023dynamic,chen2023sequential,kannan2023deft,agarwal2023dexterous,lum2024dextrah,yang2024anyrotate,wang2024lessons,lin2024twisting,sievers2022learning}. Despite their success, these methods face several limitations in practical applications. For human teleoperation, a major bottleneck is the collection of high-quality demonstrations~\cite{levine2020offline, yin2024offline}. In contact-rich dexterous manipulation, it is challenging for humans to perform safe and stable object manipulation actions, often resulting in objects falling from the hand. This makes teleoperation impractical for dexterous manipulation tasks. For sim-to-real RL, challenges arise from the significant domain gap between simulation and the real world, as well as the need for highly task-specific reward specifications when training an RL agent for complex tasks. We will discuss these challenges in more detail in Section~\ref{sec:background}.

While each approach has its own set of challenges, combining their strengths offers a promising strategy to address the complexities of dexterous manipulation. Specifically, recent sim-to-real RL works~\cite{qi2023hand,yin2023rotating} have shown that it is possible to train simple dexterous in-hand object manipulation primitives (e.g. rotation) that can be transferred to a robot in the real world. This suggests that RL can be leveraged to generate a large-scale dataset of dexterous manipulation primitives, including in-hand object rotation, translation, and grasp transitions. Meanwhile, humans excel at composing these skills through teleoperation to address more challenging tasks. For example, Yin et al. have shown that they can perform in-hand reorientation by calling several rotation primitives sequentially~\cite{yin2023rotating}. However, the external inputs in these studies are limited to a few discretized commands, lacking control over low-level interactions, such as finger movements and object contact. This limitation makes it difficult to prompt existing models to generate more detailed, finger-level interaction behaviors, such as using a syringe or screwdriver. 

Motivated by these observations, in this paper, we propose a novel training framework called \mnamefull{}~(\mname{}) to address the challenges of teaching dexterous in-hand manipulation skills. Our main idea is to use a broad, multitask simulation dataset generated via RL to pretrain a generative behavior model~(\mname) that can translate a coarse motion command to safe robot actions which can maximally preserve the motion while guaranteeing safety. In real-world applications, an external policy, such as human teleoperation or an imitation policy, can be used to prompt \mname{} to execute meaningful manipulation skills. Our approach effectively decouples high-level semantic motion generation from fine-grained low-level control, serving as a foundational low-level dexterity controller.

We validate our \mname{} framework through both simulated and real-world experiments. In simulation, we demonstrate that \mname{} significantly enhances the robustness and performance of a highly perturbed noisy policy, extending its stable operation duration by 10-100 times and enabling success even when input commands are predominantly noise. In real-world scenarios, we employ human teleoperation as a proxy for high-level motion commands and test the framework on various challenging dexterous manipulation tasks involving complex hand-object interactions across a diverse set of objects. Notably, it successfully synthesizes trajectories to solve challenging tasks, such as reorienting and using syringes and screwdrivers, with human guidance, for the first time.

%% file: src/background.tex
\section{Existing Approaches: \\ Challenges and Opportunities}
\label{sec:background}
In this section, we review the challenges and opportunities with existing approaches to dexterous manipulation that motivate our work. 
\subsection{Human Teleoperation for Imitation Learning}
\textbf{Challenge:}
Dexterous manipulation via teleoperation is challenging for humans due to the following reasons:
\paragraph{Partial Observability} During in-hand manipulation, the object motion is determined by the contact dynamics between hand and object~\cite{trinkle1990planning, okamura2000overview,ji2001planning}. Successful manipulation requires perceiving and understanding contact information, such as normal force and friction, to generate appropriate torques. However, human operators face challenges in observing this information due to occlusion and limited tactile feedback. Additionally, existing discrete haptic feedback~(e.g. binary vibration) alone is often inadequate for conveying complex touch interactions and contact geometries.

\begin{figure*}[t]
    \centering
    \includegraphics[width=1.0\linewidth]{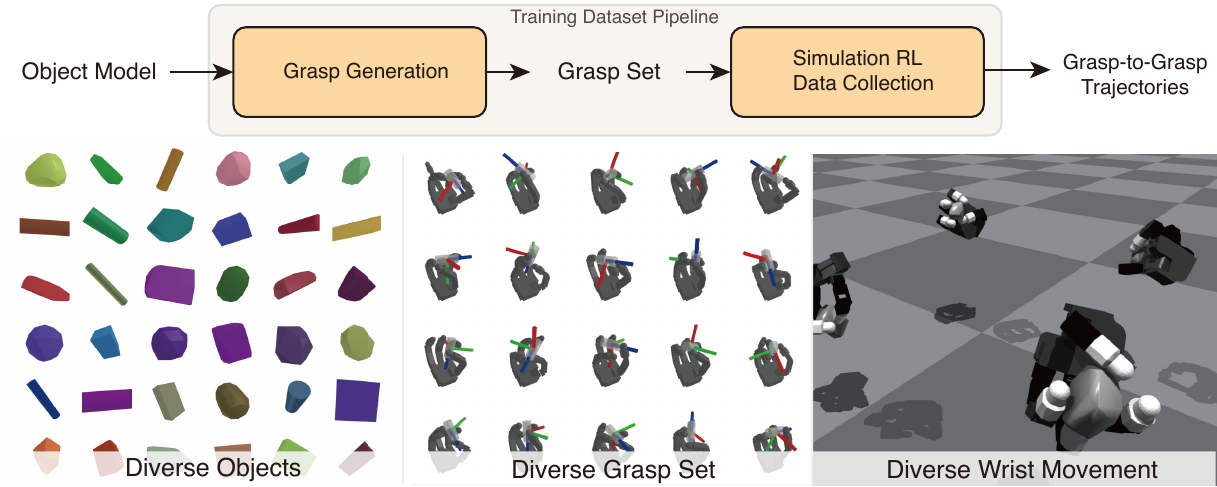}
    \caption{\textbf{Dataset:} The Anygrasp-to-Anygrasp dataset generation pipeline is designed for the generative pretraining of \mname. For a wide variety of objects, we extensively search for potential grasp configurations, using these as both the initial and goal states for RL policies. To ensure our diffusion model can manage diverse scenarios, we incorporate varied wrist poses, movements, and domain randomization during RL training and data collection.}
    \label{fig:dataset}
\end{figure*}
\paragraph{Embodiment Gap} Although human and robot hands may appear similar at first glance, they differ significantly in their kinematic structures and geometries. For example, human fingers have a smooth and compliant surfaces, while the robot fingers often have rough edges. These differences result in discrepancies in contact dynamics, making it challenging to directly transfer our understanding of human finger motions for object manipulation to robotic counterparts. In our early experiments, we find the object motion very sensitive to the change of fingertip shape.

\paragraph{Motion Complexity} Dexterous in-hand manipulation involves highly complex motion. The process requires precise control of a high degree-of-freedom dynamical system. Any suboptimal teleoperation motion at any DOF can lead to failures such as breaking grasping contacts.

\paragraph{Inaccuracy of Actions~(Force)} Existing robot hand teleoperation systems are based on hand retargeting with position control, which lacks an intuitive force control interface to users. As a result, users can only influence force through position-control errors, making teleoperation particularly challenging in force-sensitive scenarios. Moreover, the presence of noise in real world robot system further complicates control.

\textbf{Opportunity: High-level (Semantical) Motion Control} While humans may find it challenging to provide fine-grained, low-level actions directly, human teleoperation or even video demonstrations can still offer valuable coarse motion-level guidance for a variety of complex real-world tasks. Humans possess intuitive knowledge, such as where a robot hand should make contact and what constitutes a good grasp. Thus, human data can be leveraged to create a high-level semantic action plan. In locomotion research, recent studies have proposed using teleoperation commands as high-level motion prompts~\cite{cheng2024expressive}. However, extending this approach to finegrained dexterous manipulation remains an open question.

\subsection{Sim-to-real Reinforcement Learning}
\textbf{Challenge:} Developing a generalized sim-to-real policy for dexterous manipulation involves two main challenges:
\paragraph{Sim-to-Real Gap} It is difficult to reproduce real-world sensor observation~(mainly for vision input) and physics in simulation. This gap can make sim-to-real transfer highly challenging for complex tasks. In particular, transferring a vision-based control policy from simulation to real world for dexterous hands is a huge challenge and requires costly visual domain randomization~\cite{tobin2017domain}. For instance, Dextreme~\cite{handa2023dextreme} leveraged extensive visual domain randomization with 5M rendered images to train a single object rotation policy.

\paragraph{Reward Specification} A more important issue, beyond the sim-to-real gap, is the notorious challenge of designing reward functions for long-horizon, contact-rich problems. Existing methods often involve highly engineered rewards or overly complicated learning strategies~\cite{chen2023sequential}, which are task-specific and limit scalability.

\textbf{Opportunity: Low-level (Physical) Action Control} Although sim-to-real RL can be difficult, especially for those complex long-horizon or vision-based tasks, some recent works have shown that sim-to-real RL is sufficient to build diverse transferable manipulation primitives based on proprioception and touch~\cite{yin2023rotating}. Therefore, one opportunity for sim-to-real RL is to create rich low-level action primitives that can be combined with the high-level action plan discussed above. In this paper, we achieve this through generative pretraining.

%% file: src/method.tex
\begin{figure*}[t]
    \centering
    \includegraphics[width=1.0\linewidth]{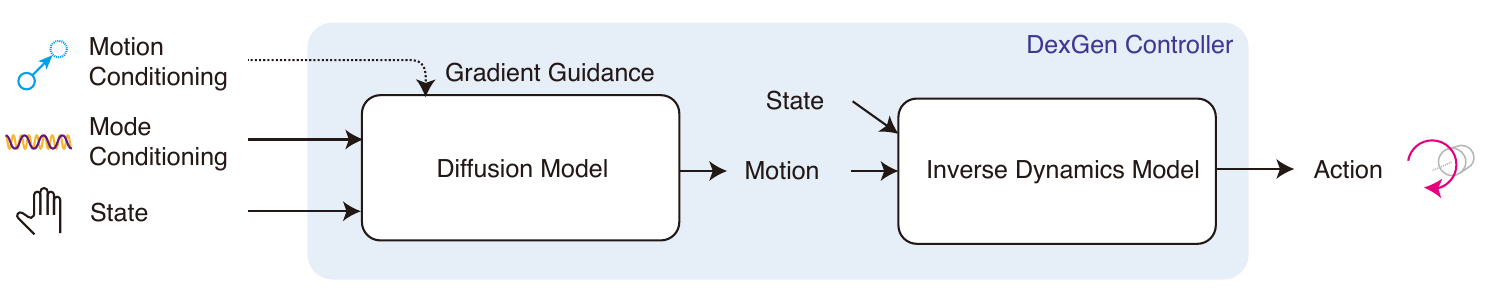}
    \caption{\textbf{Model:} Architecture of the \mname{} controller. The whole system takes robot state, external motion conditioning, and mode conditioning as input. A diffusion model first generates the motion as the intermediate action representation. The motion conditioning is not fed into the diffusion model directly but as the gradient guidance during the diffusion sampling. Then, another inverse dynamics model will translate the generated motion to executable robot action. We implement our diffusion model as a UNet in this paper. The inverse dynamics model is a residual multilayer perceptron.  }
    \label{fig:arch}
\end{figure*}
\section{The \mname{} Controller}

We propose to pretrain a generative behavior model $p_{\theta}(a|o)$ on the simulation dataset to model prior action distribution so that it can generate stable and effective actions $a$ conditioned on the robot state $o$. During inference, we can sample actions from this distribution and further aligned with external motion commands using gradient guidance. We detail the dataset used for training the model in section~\ref{ref:data}, the model architecture in section~\ref{method:arch}, and the inference procedure in section~\ref{method:inference}.

\subsection{Preliminaries}
\label{method:prelim}
\paragraph{Diffusion Models} Diffusion Model~\cite{ho2020denoising} is a powerful generative model capable of capturing highly complex probabilistic distributions, which we use as our base model. The classical form of the diffusion model is the Denoising Diffusion Probabilistic Model~(DDPM)~\cite{ho2020denoising}. DDPM defines a forward process that gradually adds noise to the data sample $x_0\sim p_{data}(x)$:
\begin{equation}
x_t = \sqrt{\alpha_t} x_{t-1} + \sqrt{1 - \alpha_t} \epsilon_t,
\end{equation}
where $\alpha_t$ is some noising schedule. We have $x_t \sim \mathcal{N}(\sqrt{\bar{\alpha}_t} x_0, (1 - \bar\alpha_t) I)$ where $\bar\alpha_t = \prod_{s=1}^t \alpha_s$ goes to 0 as $t\to+\infty$. DDPM trains a model $\mu_\theta(x_t, t)$ to predict denoised sample $x_0$ given the noised sample $x_t$ with its timestep $t$. During sampling, DDPM generates the sample by removing the noise through a reverse diffusion process:
\begin{equation}
p(x_{t-1} | x_t) = \mathcal{N}(x_{t-1}; \mu_\theta(x_t, t), \sigma_t^2 I)
\end{equation}
DDPM can generate high-fidelity samples in both vision and robotics applications. In addition to the power to generate data samples faithfully, diffusion models also support guided sampling~\cite{janner2022planning}, which turns out to be very useful in our set-up. 

Specifically,  when $p_{data}(x)$ is modeled by a diffusion model, we can sample from a product probability distribution $p(x) \propto p_{data}(x) h(x)$ where $h(x)$ is given by a differentiable energy function $h(x)=\exp J(x)$. To do this, we only need to introduce a small modification to the reverse diffusion step. Given the current sample $\mu$, we add a correction term $\alpha\Sigma\nabla J{(\mu)}$ to $\mu$. Here, $\alpha$ is a step size hyperparameter and $\Sigma$ is the variance in each diffusion step. This will guide the sample towards high-energy regions in the sample space. We can set $h(x)$ to control the style of generated samples.

\paragraph{Robot System and Notations} In this paper, we assume the robot hand is driven by a widely used PD controller. At each control timestep, we command a joint target position $\tilde{q}_t$ and the controller will use torque $\tau = K_p (\tilde{q}_t - q_t) - K_d \dot{q}_t$ to drive the joints.  Here, $q_t$ is the current joint position, and $\dot{q}_t$ is the joint velocity. $K_p$ and $K_d$ are two constant scalar gains. We use $x_t$ to denote the key point positions of finger links at time $t$ in the wrist frame. Note that our algorithm does not rely on a specific system implementation and can be extended to other robot systems. We can also specify keypoints and actions for other robots to implement our proposed algorithm.

\subsection{Large-Scale Behavior Dataset Generation}
\label{ref:data}
Since human teleoperation or external policies will control the robot hand to interact with the object in diverse ways, our model should be capable of providing refinement for all these potential scenarios~(states). To achieve this, we require a large-scale behavior dataset to pretrain our \mname{} model, ensuring comprehensive coverage of the state space. We accomplish this by collecting object manipulation trajectories in simulation through reinforcement learning. 

\begin{figure}
    \centering
    \includegraphics[width=0.99\linewidth]{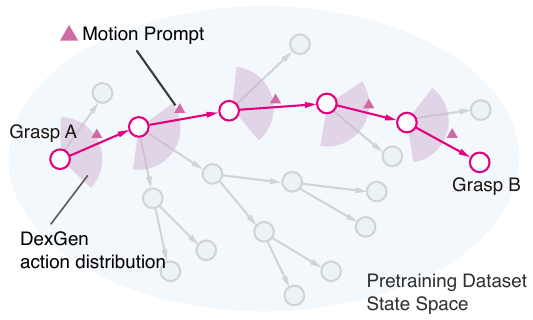}
    \caption{Our large-scale, multi-task pretraining dataset covers diverse grasp to grasp transitions~(arrows). DexGen controller learns the dataset action distribution~(purple shaded area) at each state, and we can use sequential motion prompting~(purple triangle) to perform a useful long-horizon skill, connecting two distance states.}
    \label{fig:space}
    \vspace{-0.3cm}
\end{figure}

\textbf{Anygrasp-to-Anygrasp} To ensure our dataset can cover a broad range of potential states, we introduce Anygrasp-to-Anygrasp as our central pretraining task. This task captures the essential part of in-hand manipulation, which is to move the object to arbitrary configurations. For each object, we define our training task as follows. We first generate a set of object grasps using Grasp Analysis and Rapidly-exploring Random Tree (RRT)~\cite{lavalle2001rapidly}, similar to the Manipulation RRT procedure~\cite{khandate2023sampling}. Each generated grasp is defined as a tuple (hand joint position, object pose). In each RL rollout, we initialize the object in the hand with a random grasp. We set the goal to be a randomly selected nearby grasp using the k Nearest-Neighbor search. After reaching the current grasp goal, we update the goal in the same way. We find it crucial to select a nearby reachable goal during the training process, as learning to reach a distant grasp directly can be difficult. After training, we use this anygrasp-to-anygrasp policy to rollout grasp transition sequences to cover all the possible hand-object interaction modes. We sample over 100K grasp for most objects during grasp generation to ensure coverage. This training procedure yields a rich repertoire of useful skills, including object translation and reorientation, which the high-level policy can leverage for solving downstream tasks~(Figure~\ref{fig:space}). In addition to the Anygrasp-to-Anygrasp task, we also introduce other tasks such as free finger moving and fine-grained manipulation~(e.g. fine rotation) to handle tasks that have special precision requirements. 

During RL training, we use a diverse set of random objects and wrist poses. For each task, we include random geometrical objects with different physical properties. To enhance the robustness of our policy, we randomly adjust the wrist to different poses throughout the process, in addition to employing commonly used domain randomizations, so the policy will learn to counteract the gravity effects and exhibit prehensile manipulation behavior~(Figure~\ref{fig:dataset}). By combining all these data, the robot hand can manipulate different kinds of objects in different wrist configurations against gravity rather than being limited to manipulating a single object at a certain pose. More details of the RL training can be found in Appendix.

We collect a total of $1\times 10^{10}$ transitions as our simulation dataset, equivalent to 31.7 years of real world experience. Generating this dataset~(by rolling out trained RL policies) requires 300 GPU hours. Although the dataset is large, we hypothesize it can still be far from sufficient as the human dexterity emerges from millions of years of evolution. Nevertheless, this simulated dataset still enables reliable dexterous behavior that have not been showed before.

\subsection{\mname{} Model Architecture}
\label{method:arch}
We illustrate our \mname{} model architecture in Figure~\ref{fig:arch}. The \mname{} model has two modules. The first module is a diffusion model that characterizes the distribution of robot finger keypoint motions given current observations. Here we use 3D keypoint motions $\Delta x \in \mathbb{R}^{T\times K\times3}$ in the robot hand frame as an intermediate action representation, This representation is particularly advantageous for integrating guidance from human teleoperation. In this context, $T$ is the future horizon, $K$ is the number of finger keypoints. The second module in \mname{} is an inverse dynamics model, which converts the keypoint motions to executable robot actions~(i.e. target joint position) $a_t=\tilde{q}_t$.

We use a UNet-based~\cite{ronneberger2015u} diffusion model to fit the complex keypoint motion distribution of our multitask dataset. Our model learns to generate several future finger keypoint offsets $\Delta{x}_{i} = x_{t+i} - x_{t}$ conditioned on the robot state at timestep $t$ and a mode conditioning variable. The state is a stack of historical proprioception information. The mode conditioning variable is a one-hot vector to explicitly indicate the intention of the task. For instance, when placing an object we do not want the model to produce actions that will make the robot hold the object firmly. Without introducing a ``release object'' indicator, it is hard to prompt the hand to release the object if most of the actions in the dataset will keep the object in the palm. In our dataset, the majority of transitions are labeled with a ``default'' (unconditional) label, and only a small portion of them corresponding to specialized scenarios has a special mode label. We only use a specialized precision rotation mode label for screwdriver in our experiments. For releasing object, we find that disabling DexGen controller is sufficient in practice.

The inverse dynamics model is a simple residual multilayer perceptron that outputs a normal distribution to model the actions conditioned on the current robot state and motion command. We train both the diffusion model and inverse dynamics model with our generated simulation dataset using the standard diffusion model loss function and the MSE loss for regression respectively. We train these models with the AdamW optimizer~\cite{loshchilov2017decoupled, kingma2014adam} for 15 epochs using 96 GPUs, which takes approximately 3 days. The detailed network setup can be found in the appendix. 

\begin{figure}[t]
    \centering
    \includegraphics[width=\linewidth, height=0.65\linewidth]{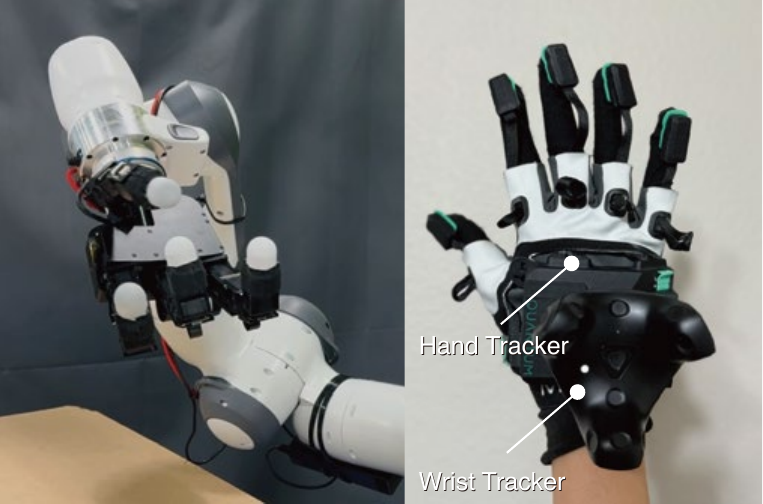}
    \caption{Real world experimental setup based on Allegro Hand with a Franka Panda Arm (Left). We use human teleoperation (Right) as a proxy for high-level policy.}
    \label{fig:enter-label}
\end{figure}

\subsection{Inference: Motion Conditioning with Guided Sampling}
\label{method:inference}
Our goal is to sample a keypoint motion that is both safe (i.e. from our learned distribution $p_\theta(\Delta x|o)$) and can maximally preserve the input reference motion. Formally, this can be written as $\Delta{x}\sim p_\theta(\Delta x|o) \exp (-{\rm Dist}(\Delta x, \Delta x_{input}))$. Here, $\Delta x_{input}\in\mathbb{R}^{K\times 3}$ is the input commanded fingertip offset, and ${\rm Dist}$ is a distance function that quantifies the distance between the predicted sequence and the input reference. There can be many ways to instantiate this distance function. In this paper, we find the following simple distance function works well empirically: 
\begin{equation}
    {\rm Dist}(\Delta{x}, \Delta x_{input}) = \sum_{i=1}^T\Vert \Delta{x}_{i} - \Delta x_{input}  \Vert^2.
\end{equation}
The above function encourages the generated future fingertip position to closely match the commanded fingertip position. Since the action of the robot hand has a high degree of freedom~(16 for the Allegro hand used in this paper), naive sampling strategies become computationally intractable. To address this, we propose using gradient guidance in the diffusion sampling process to incorporate motion conditioning. In each diffusion step, we adjust the denoised sample $\Delta{x}$ by subtracting $\alpha\Sigma\nabla_{\Delta{x}} {\rm Dist}(\Delta{x}, \Delta x_{input})$ as a guide. Here $\alpha$ is a parameter of the strength of the guidance to be tuned, which we will study in experiments. The generated finger keypoint movement is then converted to action by the inverse dynamics model. We use DDIM sampler~\cite{song2020denoising} during inference for 10Hz control. The total sampling time is around 27ms~(37Hz) on a Lambda workstation equipped with an NVIDIA RTX 4090 GPU. 

%% file: src/experiment.tex
\section{Experiments}
In the experiments, we first validate the effectiveness of \mname{} through simulated experiments, demonstrating its ability to enhance the robustness and success rate of extremely suboptimal policies. Then, we test our system in the real world with a focus on its application in shared autonomy.  Our results show that \mname{} can assist a human operator in executing unprecedented dexterous manipulation skills with remarkable generalizability.

\begin{figure}[t]
    \centering
    \includegraphics[width=1.0\linewidth,height=0.65\linewidth]{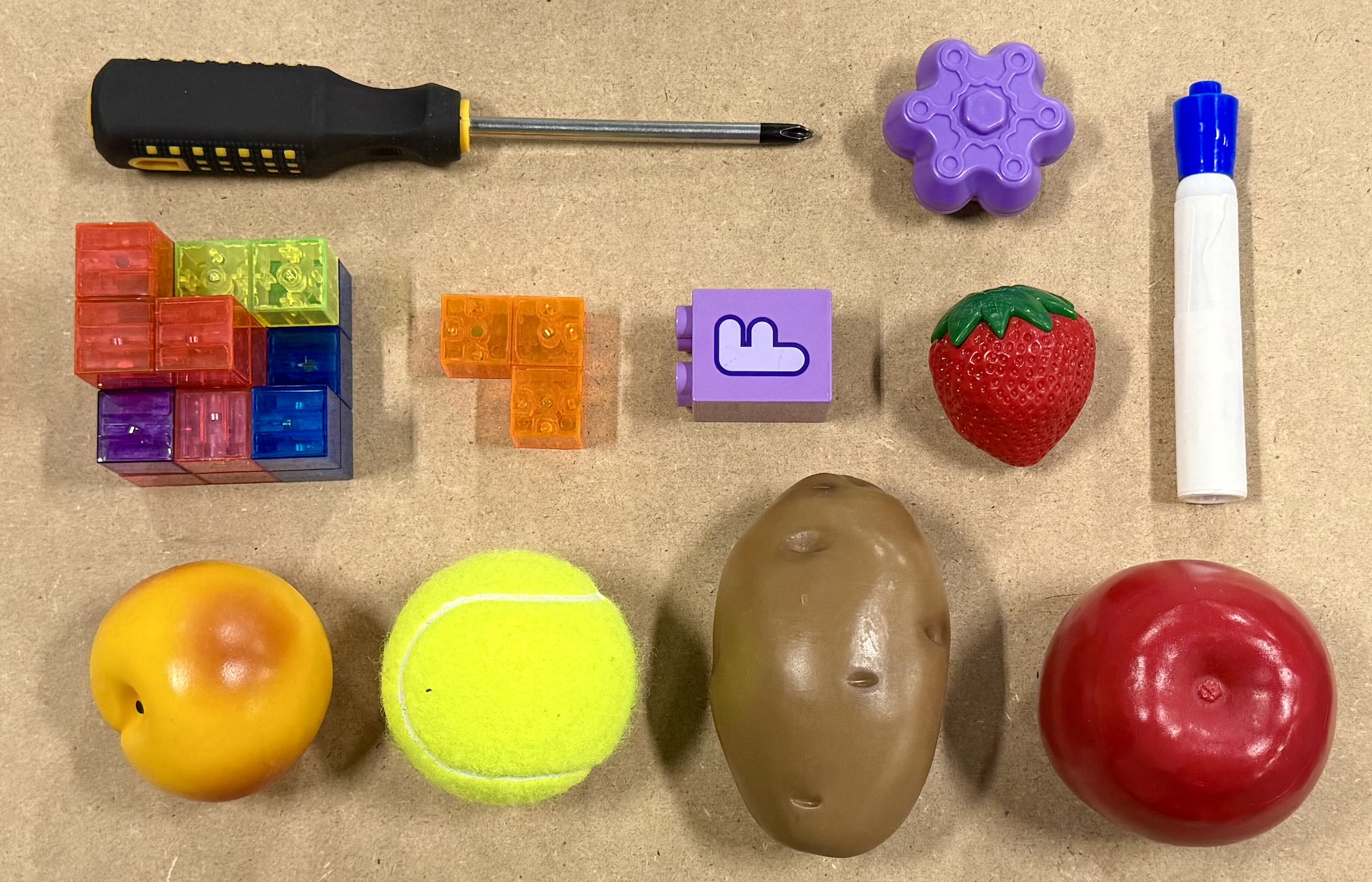}
    \caption{Part of our real world testing objects, which are not present in our pretraining dataset. We include objects of different sizes, masses, and aspect ratios.}
    \label{fig:enter-label}
\end{figure}
\subsection{System Setup}
In this paper, we use Allegro Hand as our manipulator and we attach the Allegro Hand to a Franka-panda robot arm. In the teleoperation experiments in real world, we use a retargeting-based system to control the robot with human hand gestures. The human hand pose is captured by Manus Glove and retargeted to the Allegro hand through a confidential fast retargeting method that runs at 300Hz, which we will release in a future report. We obtain the 6D human wrist pose via the Vive tracking system and use it to control the robot arm separately. Although we use this single robot setup in our experiments, we believe our method is general and can be applied to other hand setups. 
\begin{figure*}
    \centering
    \includegraphics[width=1.0\linewidth]{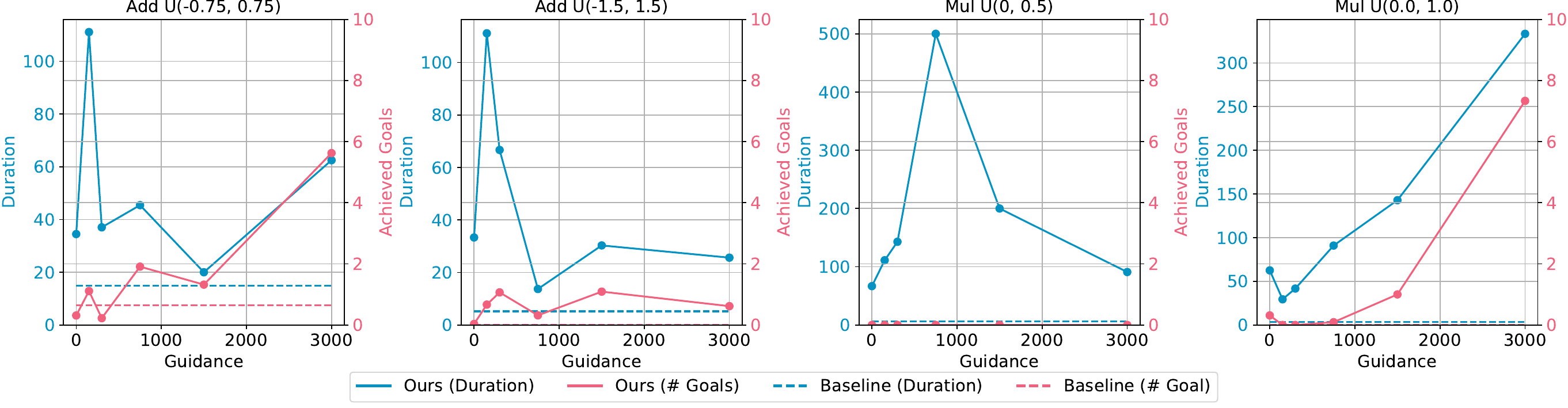}
    \caption{Results of simulation evaluation. We use \mname{} to correct several noise-corrupted expert policies. Note that each dimension of action space is bounded by [-1, 1] and these noises ruin the expert action most of the time. We measure the average duration (in seconds) and number of achieved goals per trial over a 20-minute simulated experiment. As shown in the figure, \mname{} can successfully improve the performance of these policies. Across the experiments, \mname{} can boost the duration by 10-100x and even help an extremely perturbed policy to achieve success where the baseline fails. }
    \label{fig:simulation}
\end{figure*}
\subsection{Simulated Experiments}

\subsubsection{Experimental Setup} 
We first test the capability of \mname{} in assisting suboptimal policies in solving the Anygrasp-to-Anygrasp task in simulation. We simulate 2 kinds of suboptimal policies with an expert RL policy $\pi_{exp}$. The first one is $\pi_{noisy}(a|s) = \pi_{exp}(a|s) + \mathcal{U}(-\alpha, \alpha)$, which simulates an expert that can perform dangerous suboptimal actions through additive uniform noise. The second is $\pi_{slow}(a|s) = \mathcal{U}(0, \alpha) \pi_{exp}(a|s)$, which is a slowdown version of expert. We compare these suboptimal experts $\pi$ to their assisted counterparts ${\rm \mname{}}\circ \pi$. We record the average number of critical failures~(drop the object) and the number of goal achievements within a certain time of different policies. 

\begin{figure}[t]
    \centering
    \includegraphics[width=1.0\linewidth]{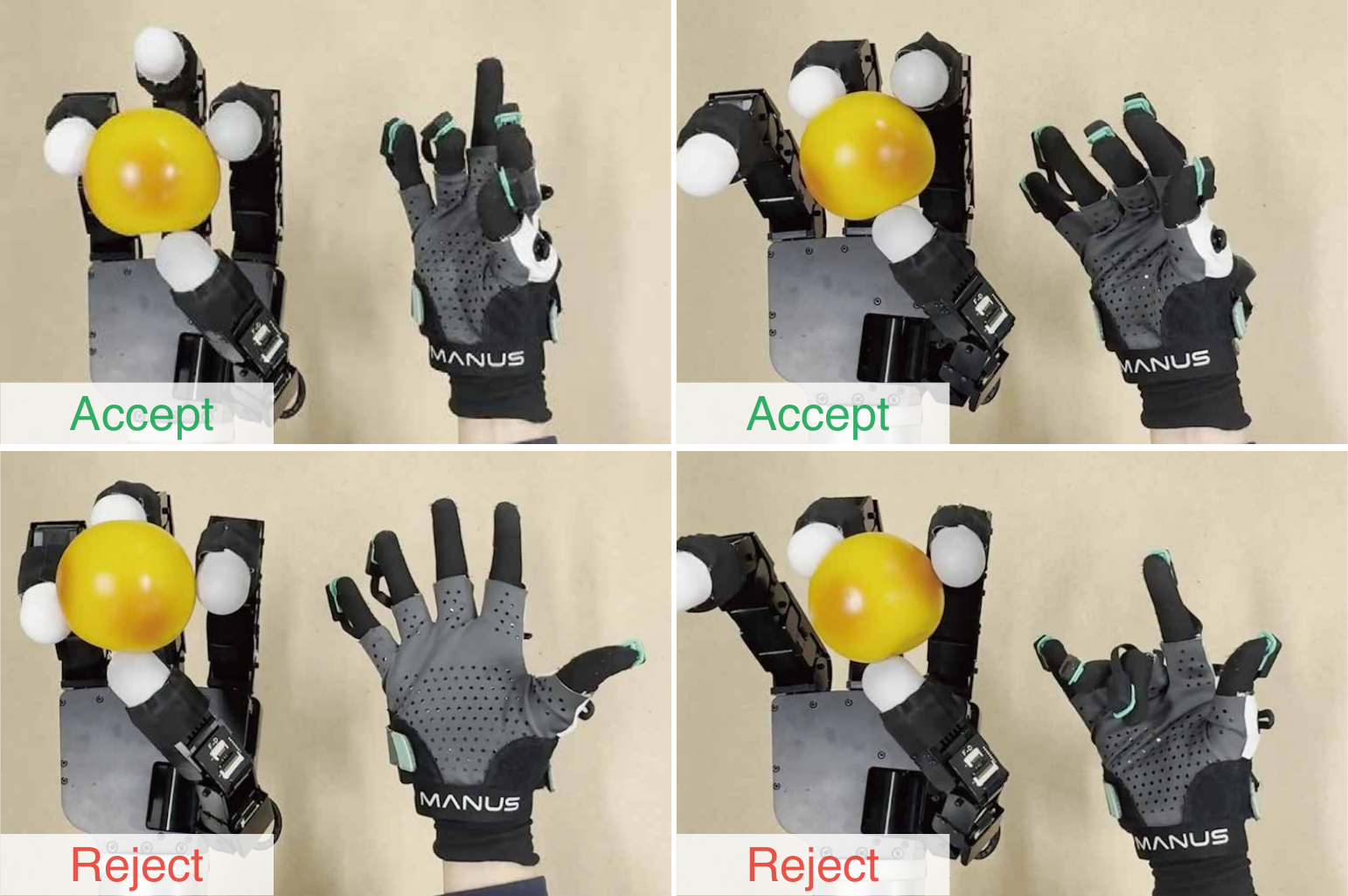}
    \caption{\mname{} can maximally preserve input action while correcting dangerous actions. \mname{} can reject users' behavior (open up the palm) and keep holding the object.}
    \label{fig:qual}
\end{figure}

\subsubsection{Main Results}  \quad We plot the result of different policies in Figure~\ref{fig:simulation}. We find that without our assistance, the noisy expert has much more frequent failures. As a result, it can only hardly achieve any goals in the evaluation. In contrast, with the assistance of \mname, we can partially recover the performance of this noisy expert. We also find that for different policies, the optimal guidance value is also different. Fortunately, there is a common region working well for all these policies. Moreover, when the guidance is relatively small, although we can maintain the object in hand, we can not achieve the desired goal as well because \mname{} does not know what the goal is. When the guidance becomes too large, the potentially suboptimal external motion command may take over \mname{} guidance and lead to a lower duration in some cases.  
\input{src/table/real}
\subsection{Real World Experiments:}
We have demonstrated that our system can provide effective assistance through simulated validation. Then, we further design several tasks for benchmarking in the real world. In the first set of experiments, we ask a human teleoperator to act as an external high-level policy and we evaluate whether our system can assist humans to solve diverse dexterous manipulation tasks. We introduce a set of atomic skills that covers common in-hand dexterous manipulation behavior.
\begin{itemize}
    \item \textbf{In-hand Object Reorientation} The user is required to control the hand to rotate a given object to a specific pose. In the beginning, we initialize the object in the air over the palm, and the user needs to first teleoperate the hand to grasp the object. 
    
    \item \textbf{Functional Grasping} Regrasping is a necessary step in tool manipulation. The user is asked to perform a power grasp on the tool handle placed either horizontally~(normal) or vertically in the air~(horizontal functional grasp). In the beginning, the user can only perform a pinch grasp or precision grasp.

    \item \textbf{In-hand Regrasping} We define this task as a harder version of object reorientation. In this task, the user is asked to achieve a specific grasp configuration~(object pose + finger pose). In the beginning, the object is initialized with a precision grasp on the fingertip.
\end{itemize}
Besides these tasks, we demonstrate some more realistic, long-horizon tasks as well. These tasks require the user to combine the skills above. In the main text, we only study the following two tasks. We leave more examples in the demo video in the appendix.
\begin{itemize}
    \item \textbf{Screwdriver} In this task, the user needs to pick up a screwdriver lying on the table and use it to tighten a bolt. 
    \item \textbf{Syringe} In this task, the user needs to pick up a syringe and inject some liquid into a target region.
\end{itemize}
\paragraph{Evaluation Protocol} We evaluate the performance of a teleoperation system by measuring the success rate a human user can achieve when using it to solve certain tasks. Before evaluation, we let users familiarize themselves with each evaluated teleoperation system in 30 minutes. Our experiments involve 2 users in this section.

\subsection{Real World Results} The performance of different approaches is shown in Table~\ref{table:real}. We observe that humans can hardly use the baseline teleoperation system to solve the tasks above. The user can drop the object easily during the contact-rich manipulation process. Compared to the baseline, our system can successfully help the user to solve many tasks in various challenging setups. During these experiments, we also observe the following intriguing properties of our system:

\paragraph{Protective ``Magnetic Effect''} We find that the fingertips show some ``magnetic effect'' when they are in contact with the object. When the user mistakenly moves a supporting finger which may drop the object, our model can override that behavior and maintain the contact as if the fingertips are sticking to the object~(Figure \ref{fig:qual} second row). This explains why the user can achieve a much higher success rate in these dexterous tasks. 

\paragraph{Intention Following} Although our model overrides dangerous user action, we find that in most cases our model can follow the user's intention~(action) well and move along the user-commanded moving direction. During the manipulation procedure, the user can still have a sense of agency over the robot hand and complete a complex task. This finding echoes our simulated result with noisy policies: DexGen can realize the intention in the noisy suboptimal actions. 

We also present a breakdown analysis of the long-horizon tasks in Table~\ref{tab:longhorizon}. For the first time, we enable such long-horizon dexterous manipulation behavior in the real world through teleoperation. Achieving tool use remains challenging as it involves several stages of complex dexterous manipulation: we can achieve a reasonable stage-wise success rate, but chaining these skills together is difficult. However, we believe that improving stage-wise policy in the future can eventually close the gap~(see the conclusion part).

%% file: src/table/real.tex
\begin{table*}

\centering
\captionof{table}{Performance of evaluated methods on the real-world tasks. We report success rate~(SR) and time-to-fall~(TTF) / Holding Time metric which is normalized by the test episode length. The raw teleoperation baseline fails completely on those tasks, while our method can help the teleoperation policy to achieve both stability and success in diverse setups. } 
\normalsize
\renewcommand\arraystretch{1.1}
\setlength\tabcolsep{0.1pt}
 \begin{tabular*}{\textwidth}{l@{\extracolsep{\fill}}cccccccc}
\toprule
\textbf{Task} &\multicolumn{2}{c}{\textbf{Reorient Large (Up)}} & \multicolumn{2}{c}{\textbf{Reorient Small (Up)}} & \multicolumn{2}{c}{\textbf{Reorient Large (Down)}} & \multicolumn{2}{c}{\textbf{Reorient Small (Down)}} \\
& SR$(\uparrow)$ & TTF$(\uparrow)$    & SR$(\uparrow)$ & TTF$(\uparrow)$   & SR$(\uparrow)$ & TTF$(\uparrow)$   & SR$(\uparrow)$ & TTF$(\uparrow)$       \\
\hline

Teleop & 0/20 & $<$5.0\% & 0/20  & $<$5.0\%  & 0/20  & $<$5.0\%  & 0/20  & $<$5.0\%     \\
Teleop + \mname & \textbf{12/20}  & \textbf{75\%} & \textbf{13/20} & \textbf{79\%}  & \textbf{10/20} & \textbf{63\%}  & \textbf{9/20}  & \textbf{58\%}     \\
\bottomrule
\toprule
\textbf{Task} &\multicolumn{2}{c}{\textbf{Func Grasp}} & \multicolumn{2}{c}{\textbf{Func Grasp (Horizontal)}} & \multicolumn{2}{c}{\textbf{Regrasp (Ball)}} & \multicolumn{2}{c}{\textbf{Regrasp (Cylinder)}} \\
& SR$(\uparrow)$ & TTF$(\uparrow)$    & SR$(\uparrow)$ & TTF$(\uparrow)$   & SR$(\uparrow)$ & TTF$(\uparrow)$   & SR$(\uparrow)$ & TTF$(\uparrow)$       \\
\hline

Teleop & 0/10 & $<$5.0\% & 1/10  & $<$10.0\% & 0/10 & $<$5.0\%  & 0/10  & $<$5.0\%     \\
Teleop + \mname & \textbf{7/10}  & \textbf{87\%} & \textbf{6/10} & \textbf{80\%}  & \textbf{5/10} & \textbf{78\%}  & \textbf{5/10}  & \textbf{74\%}     \\
\bottomrule

\end{tabular*}
\label{table:real}
\end{table*}

%% file: src/related_work.tex
\begin{table}
    \centering
    \normalsize
    \renewcommand\arraystretch{1.0}

    \caption{The breakdown success analysis of syringe and screwdriver teleoperation. These long-horizon tasks require several stages of manipulation and remain challenging.}
    \begin{tabular*}{\linewidth}{l@{\extracolsep{\fill}}|cccc}
        \toprule
       \multirow{2}{*}{\textbf{Screwdriver}\quad} & \textbf{Reorient}  &  \textbf{Regrasp} &\textbf{Align}   & \textbf{Use}\\
        \cmidrule{2-5}
       & 16/20  & 11/20 & 5/20 & 3/20\\
       \midrule
      \multirow{2}{*}{\textbf{Syringe}\quad} & \textbf{Reorient}  &  \textbf{Regrasp} &  & \textbf{Use}\\
        \cmidrule{2-5}
       & 15/20  & 9/20 &   & 4/20\\
        \bottomrule
    \end{tabular*}

    \label{tab:longhorizon}
\end{table}
\section{Related Works}

\paragraph{Foundation Models and Pretraining for Robotics}
In recent years, the success of large foundation models in natural language processing and computer vision~\cite{achiam2023gpt,kirillov2023segment,touvron2023llama} has attracted much attention in building foundation models for robotics~\cite{brohan2022rt,brohan2023rt,du2024learning,radosavovic2023robot,team2024octo,o2023open, zhao2024aloha, kim2024openvla, khazatsky2024droid}. Existing works typically focus on building a large end-to-end control model by pretraining them on large real world datasets. Our framework also leverages large-scale pretraining, but it differentiates from these works in various aspects. First, we consider pretraining a controller on pure simulation datasets rather than real world datasets which require extensive human efforts in data collection with teleoperation. Second, we study dexterous manipulation with a high DOF robotic hand and demonstrate the advantage of generative pretraining in this challenging scenario for the first time, while existing works typically consider parallel jaw gripper problems. Third, we build a low-level foundation controller that can be prompted with continuous fine-grained guidance to provide useful actions, which can be potentially integrated with high-level planning policies in the future. Most existing robotic foundation models are conditioned on discrete language prompts or task embeddings. In summary, our pretraining framework for building a foundational low-level controller presents a new perspective in the foundation model literature.

\paragraph{Shared Autonomy} Our system is also related to shared autonomy research~\cite{aarno2005adaptive, kim2006continuous,carlson2008human, schroer2015autonomous,schwarting2017parallel}, which focuses on leveraging external action guidance to produce effective actions. Some works focus on how to train RL agents with external actions~(e.g. from teleoperation)~\cite{reddy2018shared,du2020ave,reddy2022first}. In their setup, the external inputs are usually treated as part of observation fed to the RL policy. Compared to this line of work, our method does not involve human actions in the training phase. Another line of work assumes the existence of a few task-specific intentions and goals and reduces the shared autonomy problem to the goal or intent inference~\cite{aarno2005adaptive,javdani2015shared}. A limitation of this line of work in dexterous manipulation is that they do not allow fine-grained finger control since they only provide a few options for high-dimensional action space. Our method samples fine-grained low-level behavior according to user commands in high-dimensional action space and does not suffer from this problem.  The most relevant works are ~\cite{broad2019highly, yoneda2023noise}, which also use some sampled distribution to correct user behavior. We use a different correction procedure and investigate a more general and challenging dexterous manipulation setup. 

%% file: src/conclusion.tex
\section{Conclusion}
In this paper, we have presented \mnamefull{} as an initial attempt towards building a foundational low-level controller for dexterous manipulation. We have demonstrated that generative pretraining on diverse multi-task simulated trajectories yield a powerful generative controller that can translate coarse motion prompts to effective low-level actions. Combined with external high-level policy, our controller exhibits unprecedented dexterity. We believe that our work opens up new possibilities in various dimensions of dexterous manipulation. 

\textbf{Limitations and Future Work} Our exploration still has some limitations to be addressed in future works, which we discuss as follows.
\begin{enumerate}
    \item \textbf{Touch Sensing} In this work, we rely on joint angle proprioception for implicit touch sensing~(i.e. inferring force by reading control errors), which can be insufficient and nonrobust for fine-grained problems. In many cases, it is impossible to recover contact geometry based on joint angle error. In the future, we will add touch to pretraining, which has been shown possible for sim-to-real transfer. We hope that this can further improve the robustness of our system. 
    \item \textbf{Vision: Hand-Eye Coordination} Our low-level controller does not involve vision. Nevertheless, we observe that vision feedback is necessary for producing accurate tool motions for many tasks such as using a screwdriver. It is unclear whether this vision processing should be in the high-level policy or part of the proposed foundation low-level controller, and we leave this to future research.
     \item \textbf{Real-world Finetuning} In this work, we deploy our controller in a zero-shot manner. However, due to the sim-to-real gap, it can still be necessary to fine-tune our controller with some real world experience.

\end{enumerate}

%% file: src/appendix.tex
\subsection{\mname{} Training Pipeline}
We provide an overview of the full training process in Algorithm~\ref{alg:full}. The algorithm has two stages. In the first stage, we first collect manipulation trajectories with multiple RL policies. In the second stage, we distill the experience into our controller. For the dataset filtering step, we apply a very simple heuristic rule. If a rollout ends with dropping the object, then we directly discard the last $2$ second transitions. 

\begin{algorithm}[htbp]
\caption{Training Procedure of \mname{} Controller $p_\theta$}\label{alg:full}
\begin{algorithmic}[1]
\REQUIRE Manipulation tasks $\{T_i\}$ in simulation (e.g. Anygrasp to Anygrasp).
\STATE Train RL policy $\pi_i$ on each $\{T_i\}$ to convergence.
\STATE Collect training dataset $\mathcal{D}=\cup_i \small {\rm Rollout(\pi_i)}$. 
\STATE Preprocess dataset $\mathcal{D}$ by filtering failure transitions.
\STATE Train \mname{} controller $p_\theta$ on $\mathcal{D}$.
\RETURN $p_\theta$
\end{algorithmic}
\end{algorithm}

\subsection{Implementation of Anygrasp-to-Anygrasp}
The core dexterous manipulation task used by Algorithm~\ref{alg:full} is Anygrasp-to-Anygrasp. We describe its implementation as follows. 

\textbf{Grasp Generation} To define this task, we first need to generate the grasp set for each object with the Grasp Generation Algorithm~\ref{alg:graspgen}. The algorithm first generates a base grasp set using heuristic sampling, and we further expand this grasp set via RRT search to ensure that it can cover as many configurations as possible. Note that there exist many approaches for synthesizing grasps. Here, we just provide one option that works well empirically.

\begin{algorithm}[htbp]
\caption{Grasp Generation}\label{alg:graspgen}
\begin{algorithmic}[1]
\REQUIRE Object mesh $\mathcal{M}$. Initial Grasp Set Size $N$. RRT Step $N_{RRT}$.
\STATE Grasp Set $S\leftarrow {\rm HeuristicSample}(\mathcal{M}, N)$.
\STATE $S\leftarrow {\rm GraspRRTExpand} (S, \mathcal{M}, N_{RRT})$.
\RETURN $S$
\end{algorithmic}
\end{algorithm}

\begin{algorithm}[htbp]
\caption{HeuristicSample}\label{alg:hsample}
\begin{algorithmic}[1]
\REQUIRE Object mesh $\mathcal{M}$, Num Samples $N$.
\STATE $S\leftarrow [\quad]$.
\WHILE{${\rm len}(S) < N$}
\STATE $N_{pts}=\rm random([2,3,4])$.   \COMMENT{Num grasp point.}
\STATE Point $P$, Normal $n$ $\leftarrow {\rm SampleSurface}(\mathcal{M}, N_{pts})$.
\IF{${\rm GraspAnalysis}(P, n)$}
    \STATE Object Pose $p\leftarrow {\rm RandomPose()}$.
    \STATE Finger Configuration $q\leftarrow {\rm Assign}(\mathcal{M}, P, n, q)$.
    \IF{${\rm NoCollision}(q, p, \mathcal{M})$}
    \STATE $S\leftarrow S\cup \{(q, p)\}$
    \ENDIF 
\ENDIF
\ENDWHILE
\RETURN $S$
\end{algorithmic}
\end{algorithm}

\begin{algorithm}[htbp]
\caption{GraspAnalysis}\label{alg:analysis}
\begin{algorithmic}[1]
\REQUIRE Contact Points $P$, Contact Normals $n$.
\STATE $F_{min}\leftarrow$ Min solution to Net Force Opt. $(n)$.
\IF{$F_{min} < F_{thresh}$}
\STATE \textbf{return}  TRUE
\ENDIF 
\RETURN FALSE
\end{algorithmic}
\end{algorithm}

\begin{algorithm}[t]
\caption{GraspRRTExpand}\label{alg:analysis}
\begin{algorithmic}[1]
\REQUIRE Grasp set $S$, Object Mesh $\mathcal{M}$, RRT Step $N_{RRT}$.
\FOR{$i=1,2,..., N_{RRT}$}
\STATE $(q, p)\leftarrow {\rm RandomSample()}$. \COMMENT{$q$ finger configuration. $p$ object pose.}
\STATE $(q^*, p^*)\leftarrow {\rm NearestNeighbor}((q,p), S)$.
\STATE $(q, p)\leftarrow{\rm Interpolate}((q,p), (q^*, p^*))$.
\STATE $(q, p)\leftarrow {\rm FixContactAndCollision}(q, p, \mathcal{M})$.
\STATE $S=S\cup \{(q, p)\}$.
\ENDFOR
\RETURN $S$
\end{algorithmic}
\end{algorithm}

For Algorithm~\ref{alg:analysis}, we originally followed the implementation proposed by~\cite{khandate2023sampling}. However, we find that minimizing the wrench can be too strict and it is not efficient for large-scale generation. Therefore, we introduce a simplified optimization problem for grasp analysis as follows, which we find effective in practice.
\begin{tcolorbox}[colback=black!5!white, colframe=black!75!black, 
    boxsep=5pt, % Controls padding inside the box
    left=5pt, % Extra margin on the left
    right=5pt, % Extra margin on the right
    top=3pt, % Extra margin on the top
    bottom=3pt, % Extra margin on the bottom
    title=Net Force Optimization $(\{n_i\})$, % Title text
    fonttitle=\bfseries, % Title font style
    coltitle=white, % Title color 
]
\[
\begin{aligned}
    &\underset{f_i}{\text{Minimize: }}  && \left\Vert\sum f_in_i\right\Vert^2 \\
    &\text{s.t. } && \forall i, f_i\geq 0, \\
    &                   && \exists i,  f_i = 1.
\end{aligned}
\]
\end{tcolorbox}
Intuitively, we apply force $f_i$ at each contact point along contact normal $n_i$ and we optimize for a nontrivial force combination~($\exists f_i=1$) that can generate a near-zero net force. If the minimizer of this problem is below a threshold, we consider this grasp as stable. Note that the second existence constraint is hard to directly parameterize as a differentiable loss function. In our implementation, we decompose this problem into several subproblems by enforcing $f_1=1, f_2=1, ..., f_n=1$ in each subproblem.

\textbf{Reward Design} The reward function for the Anygrasp-to-Anygrasp task is as follows. It is composed of three different terms, goal-related reward $r_{goal}$, style-related reward $r_{style}$, and regularization terms $r_{reg}$.
\begin{equation}
    r = w_{goal} r_{goal} + w_{style}r_{style} + w_{reg}r_{reg}.
\end{equation}
The goal-related reward term $r_{goal}$ involves target object pose and finger joint positions:
\begin{align}
    r_{goal} &= \exp (-\alpha_{pos} \Vert p_{obj}-p_{obj}^{target}\Vert^2 -\alpha_{orn} d(R_{obj}, R_{obj}^{target})) \\ & - \alpha_{hand} \Vert q - q^{target}\Vert^2 \\
    &+ \alpha_{bonus}\mathbf{1}(\text{goal achieved}). 
\end{align}
The regularization term includes the penalty on the action scale, applied torque, and work:
\begin{align}
    r_{reg} &= - \alpha_{work} |\dot{q}^T||\tau| - \alpha_{action} \Vert a\Vert^2 - \alpha_{tau}\Vert\tau\Vert^2. 
\end{align}
For the style reward, it is a penalty term on the fingertip velocity. This can elicit different manipulation styles (fast movement or slow movement). This reward term is mainly used to boost data diversity in temporal dimensions, see discussion below. 
\begin{align}
    r_{style}=\sum_i\alpha_i \Vert \dot{x}_{tip}^{i}\Vert.
\end{align}
\textbf{Goal Dynamics}
A crucial design in the Anygrasp-to-Anygrasp task is the goal dynamics. We find that when we set a goal very far away, the RL policy can usually fail to reach that goal and as a result, the RL learning process can plateau very early. Therefore, throughout the RL process, we set goals within a moderate distance to ensure effective RL learning. Specifically, when the current goal is achieved, we search for a grasp in our grasp cache whose object distance is within a certain range as our next goal. We achieve this through a Nearest Neighbor search. Since NN search is computationally expensive for a large grasp set, we first perform a random down-sampling at each update step before the next goal computation.

\subsection{Boosting Dataset Diversity with Diverse Rewards}
To boost the diversity of the training dataset, we use multiple reward setups to train policies of different styles and use all of these policies for data collection. In this paper, we train RL policies with different $w_{style}$ and $w_{reg}$ coefficients and this yields policies of both fast and slow object manipulation behavior. This ensures that real-world states, whether they are from a good policy or a suboptimal one, are effectively managed by our controller.

\subsection{RL Training Setups}
We implement all the training tasks and data collection using the IsaacGym simulator~\cite{makoviychuk2021isaac}. We use Proximal Policy Optimization~(PPO)~\cite{schulman2017proximal} as our RL algorithm. We use asymmetric actor-critic during training, where the actor only observes proprioception information and desired goal~(represented by a relative transformation from current state to goal state), while the critic network observes all the state information such as object position and velocity etc. We use MLP to parameterize both actor and critic networks, whose hidden dimensions are both $[1024, 512, 512, 256, 256]$. We use a learning rate 0.0005, batch size 8192, PPO clip value 0.2, with $\gamma=0.99$ and GAE $\tau=0.95$. We use 8192 environments in parallel. 

\subsection{Domain Randomization}
We apply extensive domain randomizations in both training and data collection. We list the randomized components in the Table~\ref{table:dr}. 

\begin{table}[!t]
\renewcommand\arraystretch{1.05}
\caption{Domain Randomization Setup}
\centering
\begin{tabular*}{0.87\linewidth}{l@{\extracolsep{\fill}}c}
\toprule
Object: Mass~(kg)             & [0.03, 0.25]    \\
Object: Friction              & [0.5, 1.2]     \\
Object: Shape                 & $\times\mathcal{U}(0.95, 1.05)$     \\
Hand: Initial Joint Noise     & [-0.05, 0.05] \\
Hand: Friction                & [0.5, 1.2]    \\
\midrule
PD Controller: P Gain         &  $\times\mathcal{U}(0.8, 1.1)$      \\
PD Controller: D Gain         &  $\times\mathcal{U}(0.7, 1.2)$     \\
\midrule
Random Force: Scale           & 1.0/2.0       \\
Random Force: Probability     & 0.2    \\
Random Force: Decay Coeff. and Interval & 0.99 every 0.1s     \\ 
\midrule
Joint Observation Noise (white noise) &  $+\mathcal{N}(0, 0.025)$      \\
Joint Observation Noise (episode noise)      & $+\mathcal{N}(0, 0.005)$  \\
Action Noise                 & $+\mathcal{N}(0, 0.05)$   \\
\bottomrule
\end{tabular*}
\label{table:dr}
\end{table}

\begin{figure*}[t]
    \centering
    \includegraphics[width=1.0\linewidth]{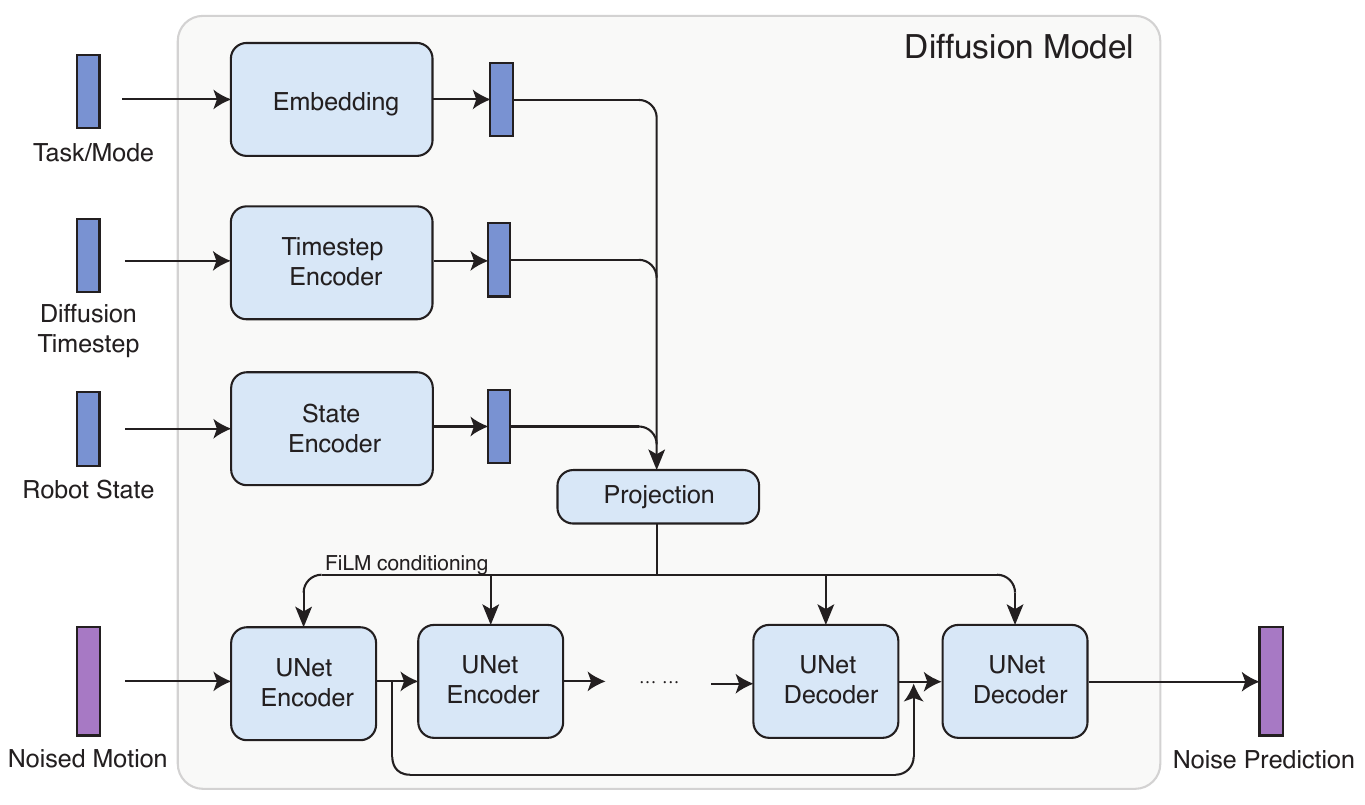}
    \caption{Diffusion Model in \mname{} Controller. We use a standard U-Net based diffusion model with FiLM conditioning. }
    \label{fig:dm}
\end{figure*}
\subsection{Diffusion Model Architecture}
We illustrate our diffusion model architecture in Figure~\ref{fig:dm}. We first use a state encoder and a mode encoder to produce a compact representation for the conditional inputs. Then, we use a UNet to predict the noise added to the current sample, with FiLM-conditioning~\cite{perez2018film} in the middle layers. We use 3 blocks for both UNet encoder and decoder. For the UNet, we use a hidden dimension of 768 and replace 1D convolution layers with fully connected layers. We implement the state encoder as a 6-layer MLP with hidden dimension 1024. We also use GroupNorm after each MLP layer with group size 8. We experimented with 8 and 12 DDIM steps during the diffusion model inference. We find there is a tradeoff between sample fidelity~(action accuracy) and latency, and they affect user experience in different ways.

In this paper, we use $T=2$ as the future motion prediction horizon, which corresponds to 0.2s future. We use $K=8$ finger keypoints~(PIP of each finger and the fingertips). In early experiments, we used $K=4$ finger keypoints (fingertips only), but this representation behaves suboptimally in the experiments and has a large inverse dynamics training loss. The simplified representation does not encode the full action information, as $K=4$ 3D keypoints only span a 12-dimensional space but the hand is 16DOF. We stack 4 history steps of robot proprioception including fingertip position, joint position, target joint position, and control error as input to the diffusion model.